# TRON: Transformer Neural Network Acceleration with Non-Coherent Silicon Photonics

Salma Afifi, Febin Sunny, Mahdi Nikdast, Sudeep Pasricha
Department of Electrical and Computer Engineering,
Colorado State University, Fort Collins, CO,
{salma.afifi, febin.sunny, mahdi.nikdast, sudeep}@colostate.edu

## ABSTRACT
Transformer neural networks are rapidly being integrated into state-of-the-art solutions for natural language processing (NLP) and computer vision. However, the complex structure of these models creates challenges for accelerating their execution on conventional electronic platforms. We propose the first silicon photonic hardware neural network accelerator called TRON for transformer-based models such as BERT, and Vision Transformers. Our analysis demonstrates that TRON exhibits at least 14× better throughput and 8× better energy efficiency, in comparison to state-of-the-art transformer accelerators.

## 1. INTRODUCTION

Transformer neural networks have gained significant popularity in the last few years, surpassing the performance of traditional Recurrent Neural Networks (RNNs) and Convolutional Neural Networks (CNNs) [1]. As the network architecture in transformer models relies on attention mechanisms and positional encodings instead of recurrence, it enables much higher parallelization than RNNs for sequence modeling and transduction problems. Since the introduction of the first transformer in 2017 [2], considerable progress has been made, with the emergence of powerful transformer-based pre-trained natural language processing (NLP) models, such as BERT [3] and Albert [4], and computer vision models, such as the Vision Transformer [5].

Despite the remarkable success of the transformer model, its size, number of parameters, and operations still require significant computational resources, hindering its progress and usage in resource-constrained systems. This highlights the main issues with these models, which includes long inference times, large memory footprint, and low computation-to-memory ratio. Existing work on inference acceleration of conventional artificial neural networks (ANNs), mainly focuses on compute-intensive operations and optimizations at the layer-level granularity, which makes extending it to transformers—with its unique layer architecture and memory-intensive requirements—challenging.

Several transformer-centric accelerators have been proposed in recent years to overcome these challenges with transformer execution [6]-[9]. However, most of the work presented so far either focuses on accelerating a specific transformer architecture or is based on electronic components. Electronic accelerators are susceptible to the limits of the post Moore's law era, where diminishing performance improvements are being observed with technology scaling. Such limitations also present major performance and energy bottlenecks for electronic dataflows [10]. On the other hand, silicon photonics has proven its proficiency as a solution beyond high-throughput communication in the telecom and datacom domains, and it is now being considered for chip-scale communication. Moreover, CMOS-compatible silicon photonic components can be used for computations, such as matrix-vector multiplications and logic gate implementations. Accordingly, the integration of silicon photonics is now actively being considered for deep learning acceleration [11].

In this paper, we introduce *TRON*, the first silicon-photonic-based transformer accelerator that can accelerate inference of a broad family of transformer models. Our novel contributions are:

- The design of a novel transformer accelerator using non-coherent silicon photonics, with the ability to accelerate any existing variant of transformer neural network models,
- Detailed crosstalk analyses, to improve signal-to-noise ratio (SNR) and tunability for photonic microresonator (MR) banks,
- A comprehensive comparison with GPU, TPU, CPU, and state-of-the-art transformer accelerators.

The rest of the paper is organized as follows. Section 2 presents a background on transformers, ANNs, and their acceleration using silicon photonics. Section 3 describes our *TRON* architecture. Section 4 discusses the experimental setup and comparisons with other accelerators, followed by conclusions in Section 5.

## 2. BACKGROUND
### 2.1. Transformer neural network models

The attention mechanism has emerged as a prominent technique in sequence learning and NLP, where long-term memory is required. By utilizing the attention mechanism, transformers have outperformed RNNs (LSTMs, GRUs) across many NLP tasks. As shown in Fig. 1, the original transformer model [2] designed for sequence learning has two main blocks: encoder and decoder. The encoder is responsible for mapping the input sequence into an abstract continuous representation. The decoder then processes that representation and gradually produces a single output while also being fed the previous outputs. Before being sent to the encoder, each input sequence is mapped to a vector, and positional encoding is used to embed the position information of each vector in relation to the original input sequence. The processed input is then passed through to the encoder/decoder block.

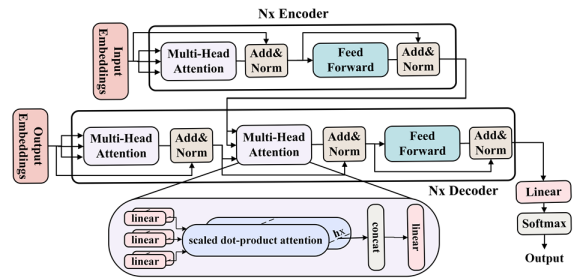

**Fig. 1. Transformer neural network architecture overview.**

The encoder and decoder blocks consist of *N* stacked layers (Fig. 1). The main sub-blocks in the encoder and decoder blocks are the multi-head attention (MHA) and feed forward (FF) layer, along with residual connections for each, followed by layer normalization. Self-attention is applied in MHA where it links each element (e.g., word) to other elements (e.g., words) in a sequence. Each MHA has *H* self-attention heads, and each attention head generates the query *(Q)*, key *(K)*, and value *(V)* vectors to compute the scaled dot-product attention. *Q, K,* and *V* vectors are generated by multiplying the MHA's input sequence *X* by the query, key, and value weight matrices: $W_Q$, $W_K$, and $W_V$. The self-attention output is then computed through a scaled dot-product operation as follows:

$$Head(X) = attention(Q,K,V) = softmax\left(QK^T/\sqrt{d_K}\right)V, \quad (1)$$

where *X* is the input matrix and $d_k$ is the dimension of *Q* and *K*. The output of the MHA is the concatenation of the self-attention heads' outputs, followed by a linear layer. The FF network is composed of two dense layers with a *RELU* activation in between.

More recent transformer-based pre-trained language models, such as BERT [3] and its variants [4], include the transformer encoder block only, as a cascaded set of *N* layers, followed by an FF layer, then *GELU*, and normalization layers. The recent Vision Transformer (ViT) model is also composed of *N* encoder layers, followed by a multi-layer perceptron [5], where the ViT's inputs are sequence vectors representing an image.

### 2.2 Transformer acceleration

Transformer accelerators in prior work focus on accelerating either a specific subset of transformer models or specific transformer layers. For instance, [7] proposed an FPGA-based hardware accelerator, for accelerating MHA and FF layers. Their approach involves efficiently partitioning the weight matrices used in the MHA and FF layers to allow both layers to share hardware resources. In [9], another FPGA-based acceleration framework was proposed with a pruning technique and a method for storing the sparse matrices. An in-memory computing-based transformer accelerator called TransPIM was presented in [6], with a novel token-based dataflow for optimized data movements along with hardware modifications to high bandwidth memory. The work in [8] proposed an automated framework called VAQF that guides the quantization and FPGA resource mapping for ViTs. Unlike prior efforts, our proposed TRON architecture can accelerate a broad family of transformer models for NLP and computer vision tasks.

### 2.3. Silicon photonics for ANN acceleration

Due to the significant benefits offered by optical ANN accelerators in terms of performance and energy efficiency, they have garnered a lot of traction from academic and industry researchers [11]. Optical ANN accelerators are either coherent or non-coherent. In coherent architectures, which use a single wavelength, parameters are imprinted onto the optical signal's phase [12] to perform multiply and accumulate (MAC) operations. Non-coherent architectures leverage multiple wavelengths and imprint parameters onto the optical signal's amplitude. Each wavelength can be used to perform operations in parallel. Current research in optical ANN accelerators has focused mainly on CNNs, MLPs, and RNNs [13]. To the best of our knowledge, *TRON* is the first optical accelerator for transformer ANN models.

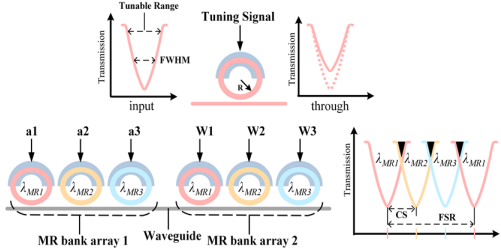

**Fig. 2.** Top microring resonator (MR) shows input and through ports' wavelengths after imprinting a parameter onto the signal. Bottom MR bank arrays perform multiplication by imprinting input activations ($a_1$-$a_3$), followed by weight vector values ($W_1$-$W_3$).

*TRON* is a non-coherent optical accelerator that uses MR opto-electronic devices (see Fig. 2) for carrying out key operations. Each MR can be designed and tuned to work at a specific wavelength, called MR resonant wavelength ($\lambda_{MR}$), defined as:

$$\lambda_{MR} = \frac{2\pi R}{m} n_{eff}, \qquad (2)$$

where $R$ is the MR radius, $m$ is the order of the resonance, and $n_{eff}$ is the effective index of the device. By carefully altering $n_{eff}$ with a tuning circuit, we can modulate electronic data onto an optical signal passing by (in the vicinity of) an MR. The tuning circuit used is either based on either thermo-optic (TO) [14] or carrier injection electro-optic (EO) tuning [15]. Both would result in a change in $n_{eff}$, and hence a resonant shift of $\Delta\lambda_{MR}$ in the MR. In non-coherent networks, computations and, specifically multiplications, are done by tuning an MR's $\Delta\lambda_{MR}$, resulting in a predictable change in the optical signal's wavelength amplitude.

To increase throughput and mimic neurons in ANNs, non-coherent architectures make use of wavelength-division multiplexing (WDM). This entails having multiple optical signals with different wavelengths in a single waveguide using an optical multiplexer [11]. The waveguide layout would pass in the vicinity of a bank of MRs, each tuned to a certain wavelength in the waveguide, to enable performing several multiplications in parallel. Fig. 2 illustrates an example of multiplying an input vector [$a_1$, $a_2$, $a_3$] by a weight vector [$W_1$, $W_2$, $W_3$]. Two MR bank arrays are used: the first imprints input activations onto the optical signals and the second performs the multiplication. The dot product output can thus be calculated by summing the three signals in the waveguide, which can be done by a photodetector (PD) device.

## 3. *TRON* HARDWARE ACCELERATOR

Our proposed *TRON* architecture is a non-coherent photonic accelerator that can accelerate the inference of a broad family of transformer models. An overview of the architecture is shown in Fig. 3. The photonic accelerator core is composed of MHA and FF units. Such composition allows reuse of resources for the encoder and decoder blocks. Interfacing with the main memory, buffering of the intermediate results, and mapping the matrices to the photonic architecture, are handled by an integrated electronic-control unit (ECU). The following subsections describe the *TRON* architecture and the hardware optimizations we have considered to efficiently accelerate transformer ANN models.

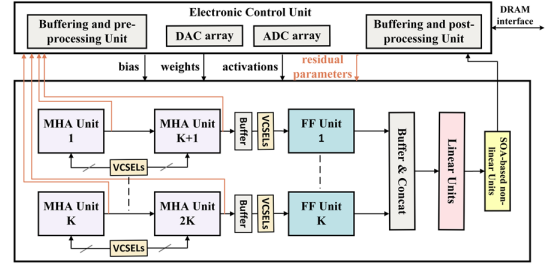

**Fig. 3.** Overview of the proposed *TRON* accelerator architecture.

### 3.1. MR tuning circuit design

MR devices in non-coherent architectures require a tuning mechanism, based on EO or TO, as mentioned earlier. In *TRON*, we employ a hybrid tuning circuit where both TO and EO are used to induce $\Delta\lambda_{MR}$. This enables us to combine the advantages of both while overcoming their disadvantages. EO tuning is faster ($\approx$ns range) and requires less power ($\approx$4 μW/nm), but it cannot be used for large tuning ranges [15]. Conversely, TO tuning accommodates a larger tunability range but at the expense of higher latency ($\approx$μs range) and power ($\approx$27 mW/*FSR*) [14]. Accordingly, in our design, EO tuning is adopted for fast induction of small $\Delta\lambda_{MR}$ in MRs, while slower TO tuning is used only when larger $\Delta\lambda_{MR}$ is required. The effectiveness of this hybrid approach was previously demonstrated in [16]. To further reduce the power overhead of TO tuning, we adopt thermal eigen decomposition method (TED) from [17]. TED entails tuning all MRs within a bank array together, which reduces power consumption. Moreover, the approach uses microheaters to perform thermal tuning which reduces thermal crosstalk noise from heat dissipated from adjoining TO circuits.

### 3.2. MR bank design-space analysis

To ensure error-free MAC operations in the optical domain, it is necessary to manage various sources of noise, namely thermal and crosstalk noise, which can interfere with parameter imprinting and degrade the network performance and accuracy. Our TED-based tuning mechanism alleviates the thermal noise that can arise from TO tuning. But non-coherent architectures, like *TRON*, are inherently noise prone due to multiple wavelengths propagating in the same waveguide which creates inter-channel crosstalk. In inter-channel crosstalk, a portion of the optical signal from neighboring wavelengths can leak into one another, causing signal distortion (see Fig. 2; bottom right). This phenomenon is further exacerbated with the presence of multiple MR banks in series, where multiple wavelengths can undesirably drop into an MR. With well-designed channel spacing (CS) and Q-factor in the MR, this can be managed by ensuring that the signal-to-noise ratio (SNR) is better than the detector sensitivity. The design of an MR should ensure adequate Q-factor to improve SNR. Additionally, the MR design should also possess sufficient tunable range, so that necessary parameters can be imprinted free of error. Mathematically, tunable range can be represented as 2×FWHM (full width half maximum), shown on the top left in Fig. 2. We optimize MR design for high FWHM and high SNR. For this optimization, we use the following models from [18]:

$$SNR\ (dB) = 10 \times \log_{10}(P_{signal}/P_{noise}), \quad (3)$$
$$P_{signal} = \Phi(\lambda_i, \lambda_j, Q) P_S(\lambda_i, \lambda_j), \quad (4)$$
$$P_{noise} = \sum_{i=1}^{n} \Phi(\lambda_i, \lambda_j, Q) P_S(\lambda_i, \lambda_j)(i \neq j), \quad (5)$$

where $\Phi$ is the crosstalk coefficient of the inter-channel crosstalk between neighboring channels $\lambda_i$ and $\lambda_j$, which is given by:

$$\Phi(\lambda_i, \lambda_j, Q) = \left(1 + \left(\frac{2Q(\lambda_i - \lambda_j)}{\lambda_j}\right)^2\right)^{-1}. \quad (6)$$

Here, $(\lambda_i - \lambda_j)$ represents the channel spacing CS, i.e., the spectral distance between two adjoining wavelengths. This is also an optimizable parameter within the confines of the free spectral range (FSR) we are considering. $P_S$ in (4) and (5) is the signal power of $\lambda_i$ that reaches the MR that is sensitive to $\lambda_j$, and can be defined as:

$$P_S = \psi(\lambda_i, \lambda_j) P_{in}(i), \quad (7)$$

where $P_{in}$ is input power to the waveguide, calculated by considering the detector sensitivity and the signal power loss of $\lambda_i$ before the MR with resonance wavelength $\lambda_j$ within the bank, represented by $\psi$. When an optical signal in a waveguide passes by an MR, the crosstalk induced power suppression in its power can be modeled as a through loss, which is defined as $\gamma$ times the signal power before it passes by the MR. This suppression factor $\gamma$ and hence $\psi$ can be calculated as follows:

$$\gamma(\lambda_i, \lambda_j, Q) = \left(1 + \left(\frac{2Q(\lambda_i - \lambda_j)}{\lambda_j}\right)^{-2}\right)^{-1}, \quad (8)$$

$$\psi(\lambda_i, \lambda_j) = \prod_{k=1}^{(k-1)<j} \gamma(\lambda_i, \lambda_k, Q). \quad (9)$$

For calculating FWHM, we use the following model:

$$FWHM = \frac{\lambda_{res}}{Q - factor}, \quad (10)$$

where $\lambda_{res}$ is the resonant wavelength of the MR being considered. Using these models, we can identify the optimal design space for our MR banks which can ensure high SNR and high tunable range ($R_{tune}$). We must also consider that the lowest optical power level ($P_{lpar}$) should be higher than $P_{noise}$, w.r.t. $P_{signal}$:

$$10\log_{10}\left(\frac{P_{signal}}{P_{lpar}}\right) < 10\log_{10}\left(\frac{P_{signal}}{P_{noise}}\right), \quad (11)$$

where $P_{lpar}$ can be defined in terms of $P_{signal}$ as follows:

$$P_{lpar} = \frac{P_{signal} \times R_{tune}}{N_{levels}}. \quad (12)$$

Replacing $P_{lpar}$ in (11) yields the following relation:

$$10\log_{10}\left(\frac{N_{levels}}{R_{tune}}\right) < SNR, \quad (13)$$

where $N_{levels}$ is the number of amplitude levels we need to represent across the available $R_{tune}$: for an n-bit parameter (ANN weight or bias) representation, $N_{levels}$ will be $2^n$. If positive and negative values are represented separately, as in the case with TRON, then $N_{levels}$ will be $2^{n-1}$. The relationship in (13) can be rearranged to obtain the relationship between $R_{tune}$ and $SNR$:

$$R_{tune} > N_{levels} \times 10^{-\frac{SNR}{10}} \quad (14)$$

Utilizing these models, we can identify the ideal design space for our MR banks, as discussed later in Section 4.1.

### 3.3. Multi-head Attention (MHA) unit design

The major challenge with transformer inference acceleration is the time-consuming matrix multiplications (MatMuls). Fortunately, these operations can be decomposed into vector dot-product operations as outlined for optical CNN acceleration in [16]. Looking closely at the self-attention in each head (1), the computation of MatMul ($Q.K^T$) cannot be performed until the generation and storage of $K^T$ completes. This dependency would infer significant power and latency overhead as we would first need to generate $K$ matrix ($K = XW_K$) optically, convert the output to digital domain, buffer the values, generate $K^T$, and then convert the matrix to the optical domain again to calculate the next MatMul ($Q.K^T$). Alternatively, using MatMul decomposition, we can rewrite the operation as two cascaded MatMul steps:

$$Q.K^T = Q.(X.W_K)^T = (Q.W_K^T).X^T \quad (15)$$

As shown by the top four MR bank arrays in Fig. 4(a), no intermediate buffering is thus needed to compute $Q.K^T$. The first two MR bank arrays generate $Q$, then by having $W_K^T$ and $X^T$ previously stored and used to tune the MRs in the following two MR bank arrays, we can directly get the output of (15) optically without any intermediate buffering or expensive opto-electric conversions. To further reduce the latency and power overhead, we propose including the scaling factor in (1) within the weight matrix ($W_K^T$) storage in the ECU. As such, the individual MR tuning values would be $W_{K\ i}^T/\sqrt{d_k}$, instead of having an additional MR bank array to perform the scaling operation. As the value of $d_k$ (dimension of $Q$ and $K$) is usually 64 in most transformer models, a simple 3-bit left shift circuit can efficiently handle the division.

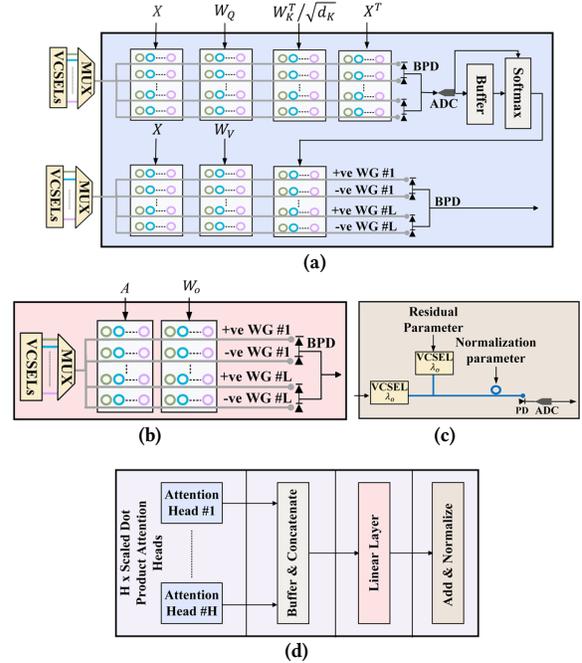

Fig. 4. (a) Attention head unit comprised of seven MR bank arrays for MatMul operations, each with dimension K×N; (b) Linear layer comprised of an MR bank array with dimension K×N; (c) Add and Normalization layers using coherent photonic summation and an MR for imprinting the normalization parameter; (d) MHA unit composed of H attention heads, buffer and concatenate block, linear layer, and an add and normalize block.

For the MatMul operations, most optical ANN accelerators (such as [13]) calculate them one-by-one, by having separate MAC units with MR bank arrays to perform the multiplication operations. Consequently, they accumulate and add the partial sums. As there are more than two consecutive MatMul operations involved in the attention computation, we avoid the accumulation of intermediate values and pass the individual multiplication results generated by the first MR bank array to the following MR bank arrays directly. The summation of all the multiplications and partial sums is then done at the end, before the softmax block, as shown in Fig. 4(a). This approach avoids the latency and power costs from early summations, intermediate buffering, and associated opto-electric conversions. Moreover, as outlined in Section 3.1, we have ensured minimal crosstalk noise, that would normally be an issue due to such MR arrangement.

Following the calculation of $(Q.K^T)$ by the upper MR bank arrays shown in Fig. 4(a), all partial sums are accumulated using balanced photodetectors (BPDs). BPDs help accommodate both positive and negative parameter values by placing separate positive and negative arms for the same waveguide. The sum acquired from the negative arm is subtracted by the BPD from the sum from the positive arm. The results are then converted to the digital domain, to undergo softmax computation.

Another challenge in MHA is the softmax operation. It is performed in each attention head and restricts parallelism as all results from the previous MatMul need to be generated first. For its implementation, we propose two optimization solutions. First, we avoid the computationally expensive division and numerical overflow by employing the log-sum-exp trick, used in a few previous works such as [7], as follows:

$$Softmax(\chi_i) = \frac{\exp(\chi_i - \chi_{max})}{\sum_{j=1}^{d_k} \exp(\chi_j - \chi_{max})}, \quad (16)$$

$$= \exp\left(\chi_i - \chi_{max} - \ln\left(\sum_{j=1}^{d_k} \exp(\chi_j - \chi_{max})\right)\right),$$

where softmax can be divided into four operations: finding $\chi_{max}$, subtraction, natural logarithm (*ln*), and exponential (*exp*). Finding $\chi_{max}$ and the subtraction can be computed using simple digital circuits. As shown in Fig. 4(a), the analog-to-digital converter (ADC) output is buffered while also being fed to a comparator circuit, so that finding $\chi_{max}$ would be computed in parallel to the MatMuls. The natural logarithm (*ln*) and exponential (*exp*) computations can be calculated using look-up tables (LUTs) [19]. This also helps get the final softmax output as an analog value from the memristor cell in the LUT, which can be used to directly tune the MR bank array. Furthermore, our scaled dot-product attention design enables high parallelism because the bottom vertical cavity surface emission laser (VCSEL) array (Fig. 4(a)) can be synchronized to only be turned on when the softmax operation is done.

The linear layer in MHA is also implemented optically using two MR bank arrays (Fig. 4(b)). For adding the MHA input to its current output (implementing the residual connection), coherent photonic summation is employed, as shown in Fig. 4(c), where the output signal from the linear layer is used to directly drive a VCSEL with wavelength $\lambda_o$. Another VCSEL with the same wavelength, is driven by value($i$) from the residual connection, and thus, when the two waveguides meet, they undergo interference, resulting in the summation of the two values. Coherent summation is ensured by using a laser phase locking mechanism [20], which guarantees that VCSEL output signals have the same phase for constructive interference to occur. Lastly, layer normalization (LN) is performed optically using a single MR, tuned by the LN parameter. The entire MHA architecture is shown in Fig. 4(d).

### 3.4. Feed Forward (FF) unit design

The FF Unit (Fig. 5(a)) is composed of two fully connected (FC) layers, with a non-linear activation in between. Each FC layer is accelerated using two MR bank arrays, with dimensions $K \times N$: one to imprint the input activations and the second to compute the MatMul between the inputs and the weight matrices. The bias values are added using coherent photonic summation, discussed in the previous section. For the non-linear unit, we implemented an optical *RELU* unit, with semiconductor-optical-amplifiers (SOAs). When the gain in an SOA is adjusted to a value close to 1, the behavior becomes almost linear, resembling the *RELU* operation. The work in [21] demonstrated how SOAs can be exploited to implement other non-linear functions such as *Sigmoid* and *tanh*. This expands the scope of *TRON* and enables us to implement the *GELU* operation (used in ViT) instead of the *RELU*, optically. The *GELU* operation can be approximated as follows [22]:

$$GELU(x) = x\Phi(x) = 0.5x(1 + \tanh\left[\sqrt{2/\Pi}\left(x + 0.044715x^3\right)\right])$$
$$= x\sigma(1.702x). \quad (17)$$

As shown in Fig 5(b), the first multiplication between 1.702 and $x$ is implemented using a single MR, and the sigmoid function is computed using the SOA implementation, described above. The last multiplication of the input with the sigmoid output is calculated using two MRs. To store the input signal and use it to tune the second MR, a low-power, local storage mechanism is used where the analog input signal from the PD is stored in a memristor cell to directly tune the last MR. The output from the non-linear unit is then buffered and used to tune the MRs in the first bank array of the second FC layer (Fig. 5(b)), to be multiplied by the weight matrix ($W_2$). Following the second FC layer, the normalization layer is implemented using an MR, the residual connection is added through coherent photonic summation, and the final normalization layer is implemented with another MR.

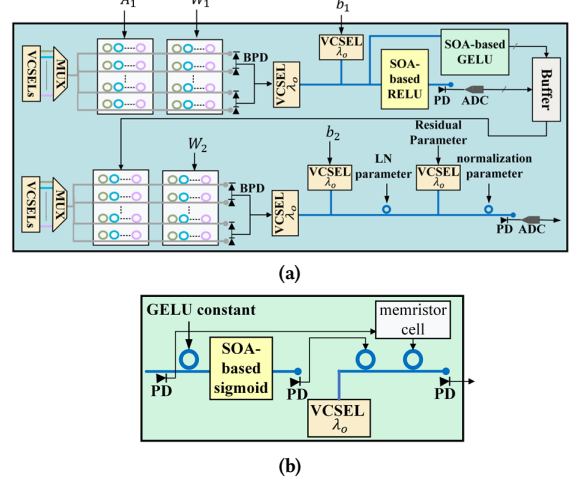

Fig. 5. (a) FF block composed of four-MR bank arrays with dimensions $K \times N$, SOA-based RELU and GELU units, and bias and residual connection additions, done with coherent photonic summation; (b) GELU unit composed of three MRs, a semiconductor-optical-amplifiers (SOA), and a VCSEL.

### 3.5. TRON architecture

The architecture of *TRON* (Fig. 3) is designed to accelerate various transformer models. The *TRON* architecture is composed of two sets of MHA units and one set of FF units. Each set has a dimension of *L*. Such an arrangement enables both the encoder and decoder blocks to easily reuse most of the units. In case of the encoder block, the first VCSEL array will be used to drive the input to the second set of MHA units only. The MHA unit can be divided into two parts: before and after the softmax operation. As softmax (see (1)) cannot be computed till the first part is completed, both parts cannot be parallelized. However, the MatMul operations in the second part can be parallelized with the MatMul operations in the FF unit. For the decoder block, the first VCSEL array is used to drive the input to the first set of MHA units. Its output is used as the input to the second MHA unit whose output then drives the FF unit. Moreover, VCSEL-reuse, as described, reduces the laser power consumption and inter-channel crosstalk. Accordingly, single VCSEL arrays are shared among rows in each MR bank array and used to imprint the input activations.

### 4. EXPERIMENTS AND RESULTS

We performed detailed simulation-based analyses to assess the efficiency of our proposed *TRON* architecture. Four transformer models were considered in our analyses: Transformer-base [2], BERT-base [3], Albert-base [4], and ViT-base [5]. The model parameters are shown in Table 1, where d*model* and d*ff* are the dimensionality of input/output and FF layers. We developed a simulator in Python to estimate the area, performance, and energy costs associated with running each model. The area, performance, and energy estimates for all electronic buffers used in *TRON* were estimated using CACTI [23] at 28nm; while the electronic circuit in softmax was synthesized using Xilinx Vivado at 28 nm and the resulting power/delay estimates were used in our analyses. Tensorflow 2.9 was used to train and analyze model accuracy.

Table 1: Transformer models and parameter counts

| Model | Params | Layers | Heads | $d_{model}$ | $d_{ff}$ |
|---|---|---|---|---|---|
| Transformer-base | 52M | 2 | 8 | 512 | 2048 |
| BERT-base | 108M | 12 | 12 | 768 | 3072 |
| Albert-base | 12M | 12 | 12 | 768 | 3072 |
| ViT-base | 86M | 12 | 12 | 768 | 3072 |

The achieved accuracies and datasets associated with each model are shown in Table 2. The Transformer, BERT, and Albert models were used for NLP tasks (language translation, sentiment analysis). ViT was evaluated using an image classification task, with pre-training on ImageNet and fine-tuning on Cifar-10. Our analysis concluded that 8-bit model quantization results in comparable accuracy to models with full (32-bit) precision (see Table 2); thus, we targeted 8-bit precision transformer models.

Table 2: Transformer model performances

| Model | Dataset(s) | Accuracy (32-bit) | Accuracy (8-bit) |
|---|---|---|---|
| Transformer-base | Ted_hrlr_translate | 66.73% | 70.4% |
| BERT-base | Sentiment-Analysis-of-IMDB-Movie-Reviews | 85.8% | 85.8% |
| Albert-base | Sentiment-Analysis-of-IMDB-Movie-Reviews | 88.3% | 88.7% |
| ViT-base | ImageNet/Cifar-10 | 97.7% | 98.0% |

The optoelectronic parameters considered for *TRON's* analysis are shown in Table 3. We considered various factors that contribute to photonic signal losses such as: waveguide propagation loss (1 dB/cm), splitter loss (0.13 dB [24]), combiner loss (0.9 dB [25]), MR through loss (0.02 dB [26]), MR modulation loss (0.72 dB [27]), EO tuning loss (6 dB/cm [15]), and TO tuning loss (27.5 mW/$FSR$ [14]). Increasing the number of wavelengths and the waveguide length will in turn increase the MR count, photonic loss, and the required laser power consumption. Accordingly, we modeled the required laser power used in our architecture for each source as:

$$P_{laser} - S_{detector} \geq P_{photo\_loss} + 10 \times \log_{10} N_\lambda, \quad (19)$$

where $P_{laser}$ is the laser power in dBm, $S_{detector}$ is the PD sensitivity in dBm, $N_\lambda$ is the number of laser sources/wavelengths, and $P_{photo\_loss}$ is the total optical loss encountered by the signal, due to the factors discussed. In the next subsection, we describe our analyses to determine the optimal values for *TRON's* architectural parameters $H$, $L$, $K$, and $N$, which were discussed in section 3.

Table 3: Parameters used for *TRON* analysis

| Devices | Latency | Power |
|---|---|---|
| EO Tuning [15] | 20 ns | 4 μW/nm |
| TO Tuning [14] | 4 μs | 27.5 mW/$FSR$ |
| VCSEL [13] | 0.07 ns | 1.3 mW |
| Photodetector [13] | 5.8 ps | 2.8 mW |
| SOA [13] | 0.3 ns | 2.2 mW |
| DAC (8 bit) [28] | 0.29 ns | 3 mW |
| ADC (8 bit) [29] | 0.82 ns | 3.1 mW |
| Memristor cell [13] | 0.1 ns | 0.07 μW |

### 4.1. TRON architecture design optimization

The *TRON* architecture design is dependent on four key parameters, as discussed in Section 3: $H$ (the number of heads in the MHA unit), $L$ (the number of layers), $K$ (the number of rows), and $N$ (the number of columns in each MR bank array). We performed an exploration to determine the optimal [$H$,$L$,$K$,$N$] configuration for *TRON*, defined as the configuration with lowest EPB/GOPS, where EPB is energy-per-bit and GOPS is giga-operations-per-second. We also set a maximum power limit of 100W for the configuration. The result of this exploration is shown in the scatterplot in Fig. 6(a). The optimal configuration [4,2,51,17], is highlighted with the pink star. This configuration is used in the comparative analysis in the following subsections. For the MR-bank design, the models described in Section 3.2 were used to perform another exploration study. Using the SNR model (3), with the $R_{tune}$ constraint (14), we explored the MR bank design space to find the parameters [$R_{tune}$,$Q$,$SNR$,$CS$], with the aim of maximizing tuning range $R_{tune}$. We considered $N_{level}$ of $2^{8-1}$, an FSR of 20 nm, Q-factor ranging from 2000 to 8000, and channel spacing ranging from 0.1 to 1 nm.

The result of the exploration is as shown in Fig. 6(b), where we have selected the data point with the best $R_{tune}$: [0.45, 6500, 24.3, 1].

### 4.2. TRON architecture component-wise analysis

To understand the performance of the major components within the *TRON* architecture, we present a breakdown in terms of power and latency for these components in Fig. 7. For the power, it is evident that MatMul operations in the attention heads contribute to more than half of the architecture's power overhead. This is because of the large dimensions of the matrices being multiplied in the MHA blocks, in each attention head. This requires many digital-to-analog converters (DACs), whose power consumption is considerable. Moreover, the sequential dependency in the attention head also contributes notably to the latency overhead. As Albert shares all attention and FF parameters across layers [4], this leads to a minimization of the number of active DACs, reducing the overall power consumption for the Albert model.

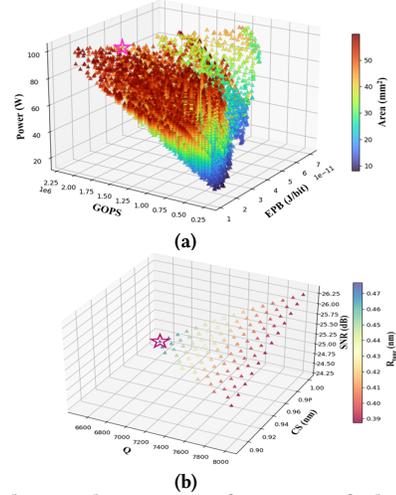

(a)

(b)

Fig. 6. (a) Architectural optimization for *TRON*, to find optimal [$H$, $L$, $K$, $N$] configuration with best energy-efficiency and throughput. The best configuration is shown with the pink star; (b) MR bank optimization for *TRON*, to identify optimal [$R_{tune}$,$Q$,$SNR$,$CS$]. The best design point with the highest $R_{tune}$, is shown with the pink star.

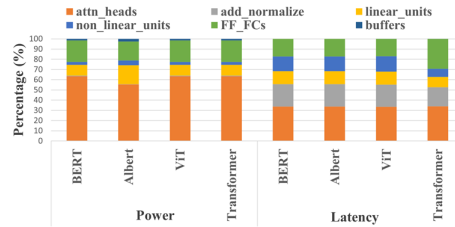

Fig. 7. Power and latency breakdown across *TRON* components.

### 4.3. Comparison to state-of-the-art accelerators

We compared *TRON* execution on multiple processors and state-of-the-art transformer accelerators: Tesla V100-SXM2 GPU, TPU v2 [30], Intel Xeon CPU, TransPIM [6], FPGA transformer accelerator in [7] (FPGA_Acc1), VAQF [8], and FPGA transformer accelerator in [9] (FPGA_Acc2). VAQF focuses on vision transformers and FPGA-Acc2 on traditional encoder-decoder transformer architectures and transformer-based language models; results for these two platforms are thus restricted to the models they are targeted for. We used power, latency, and energy values reported for the selected accelerators, and results from executing models on the GPU/CPU/TPU platforms to estimate the EPB and GOPS for each model. The *TRON* architectural configuration used in the comparisons is the one described in Section 4.1.

Fig. 8 shows the GOPS comparison between *TRON* and the other architectures considered. Our architecture achieves on average 262×, 1631×, 1930×, 14×, and 55× better GOPS than GPU, TPU, CPU, TransPIM, and FPGA_Acc1, respectively. When comparing transformer model-specific accelerators, *TRON* has on average 352× higher GOPS than FPGA_Acc2 for transformer, BERT, and Albert models, and 846× higher GOPS than VAQF for ViT. The higher throughput over all compute platforms can be explained in terms of *TRON*'s high-speed execution in the optical domain and minimal computations in the digital/electric domain.

Fig. 9 shows the energy-per-bit (EPB) comparison. On average, *TRON* attains 4231×, 12397×, 10971×, 14×, and 8× lower EPB than GPU, TPU, CPU, TransPIM, and FPGA_Acc1. For model-specific accelerators, we achieve on average 802× lower EPB than FPGA_Acc2 for transformer, BERT, and Albert models, and 32× lower EPB than VAQF for ViT. These EPB improvements can be attributed to *TRON*'s low latency operations and relatively lower power compared to some of the computation platforms considered.

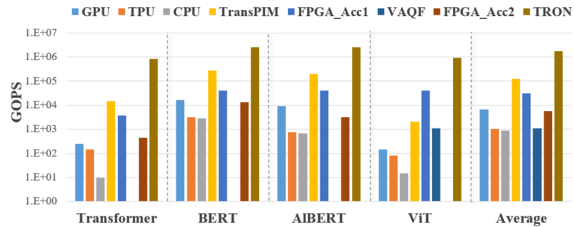

Fig. 8. Throughput comparison across transformer accelerators

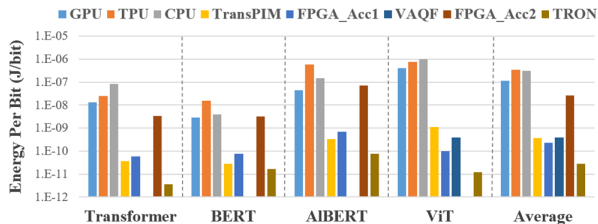

Fig. 9. EPB comparison across transformer accelerators.

### 4.4. TRON for edge environments

Edge computing environments have stringent power constraints for accelerators. We performed a design space exploration, similar to that described in Section 4.1, to find an edge-friendly *TRON* configuration with a power limit of 10W (instead of 100W that we considered earlier). We identified the optimal edge configuration values for [$H,L,K,N$] as [4,1,12,12]. Fig. 10 illustrates a comparison for the average power, GOPS, and EPB values across models, among *TRON_edge, TRON,* and the platforms previously discussed. The values shown are normalized to those obtained for the CPU. Our *TRON_edge* accelerator consumes on average, considerably lower power (~10W). While the GOPS values slightly decrease, the edge configuration's throughput still outperforms all compute and accelerator platforms by at least 16%. The EPB value for *TRON_edge* is higher than for *TRON* but it still is on average 4× to 6292× lower than all other platforms. In this manner, TRON can be customized to provide the best performance and energy-efficiency for any given target power consumption constraint.

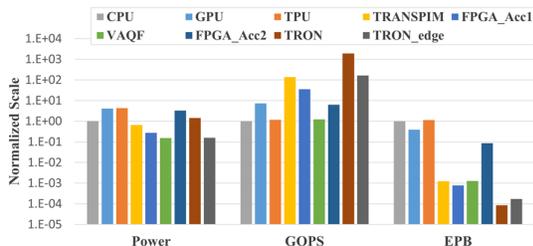

Fig. 10. *TRON*_edge's power, throughput, and energy comparison.

### 5. CONCLUSIONS

In this paper, we presented the first non-coherent silicon photonic hardware transformer accelerator, called *TRON*. Our proposed accelerator architecture exhibited throughput improvements of at least 14× and energy-efficiency improvements of at least 8× when compared to eight different processing platforms and state-of-the-art transformer accelerators. These results demonstrate the promise of *TRON* in terms of energy-efficiency and high-throughput inference acceleration for transformer neural networks. This work focused on the hardware architecture design with silicon photonics. When combined with software optimization techniques that aim to reduce a transformer's large memory footprint, significantly better throughput and energy efficiency can be achieved.